\documentclass[a4paper,twoside]{article}

\usepackage{epsfig}
\usepackage{subcaption}
\usepackage{calc}
\usepackage{amssymb}
\usepackage{amstext}
\usepackage{amsmath}
\usepackage{amsthm}
\usepackage{multicol}
\usepackage{pslatex}
\usepackage{apalike}
\usepackage{algorithm2e}
\usepackage[bottom]{footmisc}
\usepackage{booktabs}
\usepackage[table]{xcolor}  
\usepackage{colortbl}
\usepackage{hhline}
\usepackage{graphicx} 
\usepackage{float}
\usepackage{diagbox}

\usepackage{SCITEPRESS}     

\begin{document}

\title{On the exploitation of DCT statistics for cropping detectors}

\author{\authorname{Claudio Vittorio Ragaglia\sup{1}\orcidAuthor{0009-0000-1598-5226}, Francesco Guarnera\sup{1}\orcidAuthor{0000-0002-7703-3367} and Sebastiano Battiato\sup{2}\orcidAuthor{0000-0001-6127-2470}}
\affiliation{\sup{1} Department of Mathematics and Computer Science, University of Catania, Viale Andrea Doria 6, Catania, 95125, Italy}
\email{claudio.ragaglia@phd.unict.it, francesco.guarnera@unict.it, sebastiano.battiato@unict.it}
}


\abstract{The study of frequency components derived from Discrete Cosine Transform (DCT) has been widely used in image analysis. In recent years it has been observed that significant information can be extrapolated from them about the lifecycle of the image, but no study has focused on the analysis between them and the source resolution of the image. In this work, we investigated a novel image resolution classifier that employs DCT statistics with the goal to detect the original resolution of images; in particular the insight was exploited to address the challenge of identifying cropped images. Training a Machine Learning (ML) classifier on entire images (not cropped), the generated model can leverage this information to detect cropping. The results demonstrate the classifier's reliability in distinguishing between cropped and not cropped images, providing a dependable estimation of their original resolution. This advancement has significant implications for image processing applications, including digital security, authenticity verification, and visual quality analysis, by offering a new tool for detecting image manipulations and enhancing qualitative image assessment. This work opens new perspectives in the field, with potential to transform image analysis and usage across multiple domains.}

\onecolumn \maketitle \normalsize \setcounter{footnote}{0} \vfill

\section{\uppercase{Introduction}}
\label{sec:introduction}

In today's digital age, images serve as a critical medium for communication, documentation and evidence in areas ranging from journalism and social media to security and legal proceedings. The authenticity and integrity of visual content have never been more important, given the ease with which digital images can be manipulated using sophisticated editing tools. Among myriad possible alterations, cropping and resolution manipulation pose significant challenges to maintaining trust in digital images. Determining whether an image has been cropped or identifying its original resolution is critical to verifying the authenticity of digital content and protecting against misinformation. The state of the art has already showcased that the pivotal essence of a digital image resides within the domain of Discrete Cosine Transform (DCT). This information needs proper exploration to unlock its potential for image processing; such applications encompass but are not limited to object recognition, scene recognition, and image forensic analysis \cite{battiato2001psychovisual,ravi2016semantic}.The authors of \cite{LamGoodman2000} proved that although different model could be suitable to describe images, the Laplacian distribution remains a choice balancing simplicity of the model and fidelity to the empirical data, especially considering it can be described by only the $\beta$ value, a scale parameter crucial for determining the distribution's spread, as will be described in more detail in next sections. This paper introduces a new way to detect previous crop using a classifier designed to identify the source resolution, and trained with the aforementioned features. The underlying idea of this classifier is rooted in the observation that while an image's physical dimensions can be altered, the intrinsic properties encoded in its frequency domain remain indicative of its original resolution. By leveraging these properties, our classifier aims to discern the original resolution category of an image, providing a valuable tool for detecting image manipulations such as cropping. In summary, our framework seeks to exploit the distinctive patterns encapsulated in the $\beta$ values for resolution classification, offering a novel tool for the detection of cropping.
The present document details the initial findings of our investigation and it is intended as a preliminary work, which will be deeply investigated in future. At the best of our knowledge, the methodology proposed in this paper represents a novel approach, with no similar methods currently existing.

The remainder of this paper is organized as follows. Section \ref{sec:related_works} presents a brief overview of state of the art related to the use of same features in forensics field, Section \ref{sec:method} describes our idea for cropping detection, in Section \ref{sec:results} the results of our tests are reported and finally Section \ref{sec:conclusion} concludes the paper.

\section{\uppercase{Related works}}
\label{sec:related_works}

The history reconstruction of digital images has always been a topic of interest in image forensics. As described in \cite{piva2013overview}, the advent of accessible imaging technology coupled with sophisticated image editing software has increased the potential for tampering of visual content. Historically, visual data was regarded as a reliable testament to truthfulness, however the digital era has introduced scenarios that challenge this trust. In this context, digital images can be manipulated to change visual content, able to blur the line between authentic and manipulated images.
The authors of \cite{piva2013overview,battiato2009digital,4806202}  delineate the evolution of image forensics, a field dedicated to verifying the history and credibility of digital images to assess their authenticity and obtain information for forensic purposes. 


This field has witnessed rapid growth, spurred by the urgent need for tools capable of exposing the manipulations an image may have undergone throughout its lifecycle. The increasing sophistication of processing tools only underscores the importance of advancing forensic methodologies capable of keeping pace with evolving technologies. 
The detection of digital forgeries stands as a cornerstone in the field of digital forensic science, confronting the challenge of identifying unauthorized manipulations within multimedia content. This area encompasses a wide array of techniques and methodologies, each designed to unveil specific types of alterations, from simple image modifications to the creation of entirely synthetic content, such as deepfakes \cite{giudice2021fighting}. At the heart of these investigations lies the imperative to preserve the visual material's integrity and authenticity, which are pivotal in legal, and security contexts. 

The social media explosion in conjunction with the use of JPEG compression move the forensic researcher to focus on this specific scenario; for this reason during image forensic analysis one of the first task faced is the detection of double quantization (DQD), a phenomenon that occurs when a JPEG image is compressed two times, leaving traces in the frequency domain. These traces have been analyzed through statistical methods \cite{barni2017aligned,hou2013double} and through machine learning ones \cite{uricchio2017localization,10.1007/978-3-030-30645-8_65} to determine if an image has undergone post-acquisition manipulations, providing investigators with a powerful tool to identify forgeries.

First Quantization Estimation (FQE) plays a key role, allowing the deduction of the quantization matrix employed during the first quantization giving the possibility to do hypothesis about the camera model of the acquired image. This technique leverages knowledge of JPEG compression characteristics, including quantization factors, to estimate the image's original conditions. Accurately estimating these parameters \cite{galvan2014first,tondi2021boosting,battiato2022cnn} is crucial for identifying modified images, as it provides a benchmark against which the residual traces of the second compression can be compared.



\section{\uppercase{METHOD}}
\label{sec:method}

\begin{figure}[!t]
  \centering
   {\epsfig{file = 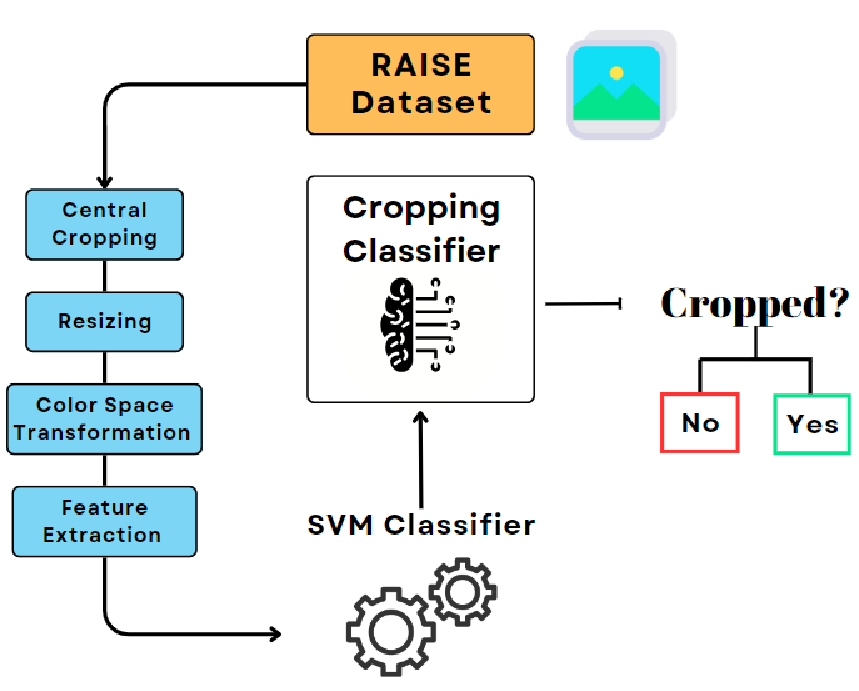, width = \linewidth}}
  \caption{Pipeline for Images Dataset and SVM Classifier Development}
  \label{fig:flowchart}
 \end{figure}

Our study focuses on refining and applying studies exploiting DCT features not only related to JPEG scenarios. By analyzing DCT coefficients and exploring new classification methods based on its distributions, we aim to develop a robust framework for precise image manipulation identification.
The significance of this work lies not only in its practical application for image security and authentication but also in its contribution to the theoretical understanding of the limits and potentials of the use of DCT distributions in digital forensic scenarios.  The flowchart in Figure \ref{fig:flowchart} shows the overall pipeline used for creating the dataset and building the SVM classifier.

\subsection{Discrete Cosine Transform}

W.r.t. Fourier Transform, which converts signals from spatial domain to the frequency ones using a combination of sine and cosine functions, DCT 
exclusively uses cosine functions. This distinction is particularly advantageous for image processing applications, where the signal (image) is real-valued. The Fourier Transform of a discrete signal $f(x,y)$ is defined as:
 
\begin{equation}\label{eq:FourierTransform}
F(k, l) = \sum_{x=0}^{N-1} \sum_{y=0}^{M-1} f(x, y) e^{-2\pi i\left(\frac{kx}{N} + \frac{ly}{M}\right)}
\end{equation}

Here, $F(k,l)$ represents the complex frequency spectrum of the image, with k and l indicating the frequency components in the horizontal and vertical dimensions, respectively. The exponential term includes both sine and cosine components, reflecting the signal's frequency content. The DCT, however, focuses on the cosine terms, which are more efficient for images due to the following reasons. The first is Energy Compaction: for many real-world images, the DCT exhibits superior energy compaction properties, meaning that a significant portion of the image's visual information tends to be concentrated in a few low-frequency components of the DCT. This characteristic is highly beneficial for image compression. The second is Real-valued Output: Unlike the Fourier Transform, which produces complex numbers, the DCT yields real numbers. This simplicity is advantageous for the storage, processing, and interpretation of the results.

In digital image processing, the Discrete Cosine Transform (DCT) is a critical technique for converting images from the spatial domain to the frequency ones. This conversion is essential for a wide range of applications, including compression, enhancement, and detailed analysis of images. DCT operates by decomposing an image into a sum of cosine functions of varying frequencies, each one representing a distinct component of the frequency spectrum. The one-dimensional Discrete Cosine Transform (DCT) is defined by Equation \ref{eq:DCT_one_dimension}. $C(u)$ is the DCT value at index $u$, defined by the equation:

\begin{equation}\label{eq:DCT_one_dimension}
C(u) = \alpha(u) \sum_{x=0}^{N-1} f(x) \cos\left[\frac{\pi(2x+1)u}{2N}\right]
\end{equation}
where $f(x)$ is the value of the input signal at position $x$, $N$ is the total number of samples in the signal or the length of the input signal, $x$ is the index of the current sample within the input signal, and \( \alpha(u) \) is a normalization factor.

To note that for 2-dimensional input as images the DCT is applied for both the axis; given the image $I$ as a matrix, the output is defined as follow:
\begin{equation}\label{eq:DCT_two_dimensions}
    DCT_{2D}(I)=DCT(DCT(I_{T})_{T})
\end{equation}
where the pedix $T$ indicates the transpose. The DCT's zero-frequency component, or the DC (Direct Current) component, reflects the average brightness across the entire image, serving as a reference for the higher-frequency AC (Alternating Current) components. These AC components encode the variations in pixel intensities, capturing the detailed textures and contours of the image. The magnitude of each AC coefficient reveals the strength of a specific frequency pattern within the image, while its phase angle indicates the pattern's spatial orientation. Utilizing DCT in image compression, such as in the JPEG standard, involves prioritizing the lower frequency components, which are more significant to human perception, and reducing the higher frequency components that contribute less to the overall visual quality. This method effectively reduces data redundancy without substantially degrading image quality. Moreover, transitioning images into the frequency domain using DCT reveals underlying patterns and relationships that are not visible in the spatial domain. This characteristic is invaluable for enhancing the efficacy of algorithms designed for tasks like image resolution classification. 
The mathematical formulation of DCT offers a solid theoretical foundation for understanding its application in digital image processing, underscoring its versatility and efficiency in encoding and analyzing visual information. 

\begin{figure}[!t]
  \centering
   {\epsfig{file = 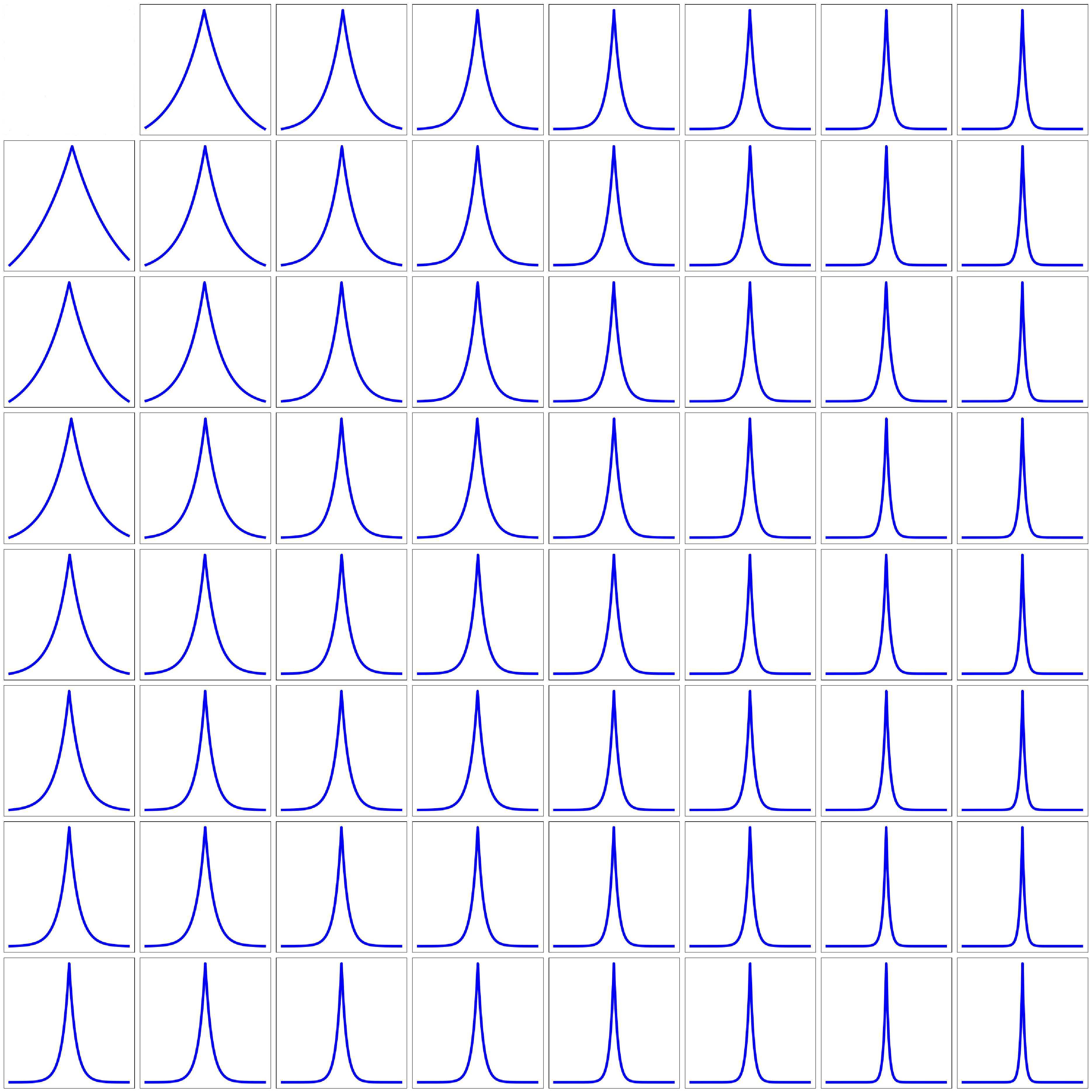, width = \linewidth}}
  \caption{Example of probability distribution of DCT coefficients, fitting a Laplacian distribution}
  \label{fig:laplacian}
 \end{figure}

A foundational piece in understanding the utility of DCT in applications of Image Processing is the work conducted by \cite{LamGoodman2000}. For the first time, Lam proposed that the AC coefficients of an image follow a distribution that can be described as Laplacian distributions (Figure \ref{fig:laplacian}, characterized by two parameters: $\mu$ and $\beta$. The first parameter ($\mu$) indicates the peak and the second ($\beta$) is the spread of the distribution, offering profound insights into the nature of image data in the frequency domain. Lam's analysis revealed that these Laplacian distributions (Figure \ref{fig:laplacian}), with their distinct parameters, could effectively model the behavior of AC coefficients, providing a mathematical framework that has since been instrumental in various image processing applications, from compression to authentication and beyond. The implications of \cite{LamGoodman2000} are far-reaching, enabling a deeper understanding of image characteristics in the frequency domain and facilitating the development of more sophisticated algorithms for image analysis, including the resolution classification approach that forms the core of our study.

\subsection{Dataset}
\label{sec:dataset}
Our study started from RAISE \cite{dang2015raise}, a comprehensive collection of 8156 high-resolution, unaltered photographic images designed for digital image forensics. The dataset is notable for its diversity, including thousands of images spanning a wide range of scenes and subjects, making it an ideal candidate for studies in image processing and manipulation detection. 
The dataset was processed in order to adapt it to our studies on image resolution classification. Therefore, we employed a specific processing pipeline to prepare the images for analysis, as follows:

\textbf{Central Cropping}. In the preprocessing phase, a central cropping was applied to each image to convert them into square dimensions. This was achieved by selecting the smaller value between height and width to define the new dimension for all sides of the images. This simplification was adopted for the preliminary study detailed in this paper.

\textbf{Resizing}. The cropped images were then resized to create five distinct datasets at resolutions of $2048\times2048$, $1024\times1024$, $512\times512$, $256\times256$, and $128\times128$ pixels. This resizing was achieved through bicubic interpolation, a method known for its effectiveness in preserving image quality during downsizing by utilizing cubic polynomials to interpolate the pixel values.

\textbf{Color Space Transformation}. In this step, images were converted to the \textit{YCbCr} color space, retaining only the luminance channel (Y). The transition to the luminance channel is a common practice in DCT analysis because the human eye is more sensitive to variations in brightness than color. Focusing on the luminance channel allows for a more efficient and relevant analysis of image details and textures in the frequency domain.

\textbf{Feature extraction}. The luminance matrix was divided in $8\times8$ blocks. We utilize 8x8 grids for feature extraction based on established findings that this size yields better results for calculating features, a phenomenon well-documented in the study of JPEG compression algorithms. For each block the DCT was carried out obtaining 64 values and then scrolling through the blocks of the mentioned values the distribution  was generated. It is easy to understand that the size of the distribution is related to the starting size of the image, but fitting every distribution to a Laplacian ones every distribution will be reconducted to a single value $\beta$, making the feature independent by the starting image resolution.

As a result of this comprehensive pipeline, we constructed a DataFrame encapsulating the $\beta$ values of the AC distributions across all images at various resolutions. This dataset forms the basis for our analysis, enabling us to develop a classifier that can discern image resolution and detect cropping.

In summary, the pipeline applied to a single image from the RAISE dataset for our study involved these steps: central cropping to square dimensions, resizing to multiple resolutions via bicubic interpolation, transforming to the YCbCr color space and retaining only the luminance channel, followed by dividing the luminance matrix into 8x8 blocks to perform DCT, from which the distribution of coefficients was fitted to a Laplacian distribution to derive a singular $\beta$ value as the feature for resolution classification and cropping detection.

\subsection{Classifier}

The primary target of our research is to develop a model capable of accurately classifying the resolution of an image based on the $\beta$ coefficients of its AC distributions. Our approach focuses on a classification system that distinguishes among $5$ specific resolution classes: $2048\times2048$, $1024\times1024$, $512\times512$, $256\times256$, and $128\times128$ pixels. These resolutions were selected to cover a broad spectrum of common image sizes, from high-resolution to smaller.

\begin{figure}[!t]
  \centering
   {\epsfig{file = 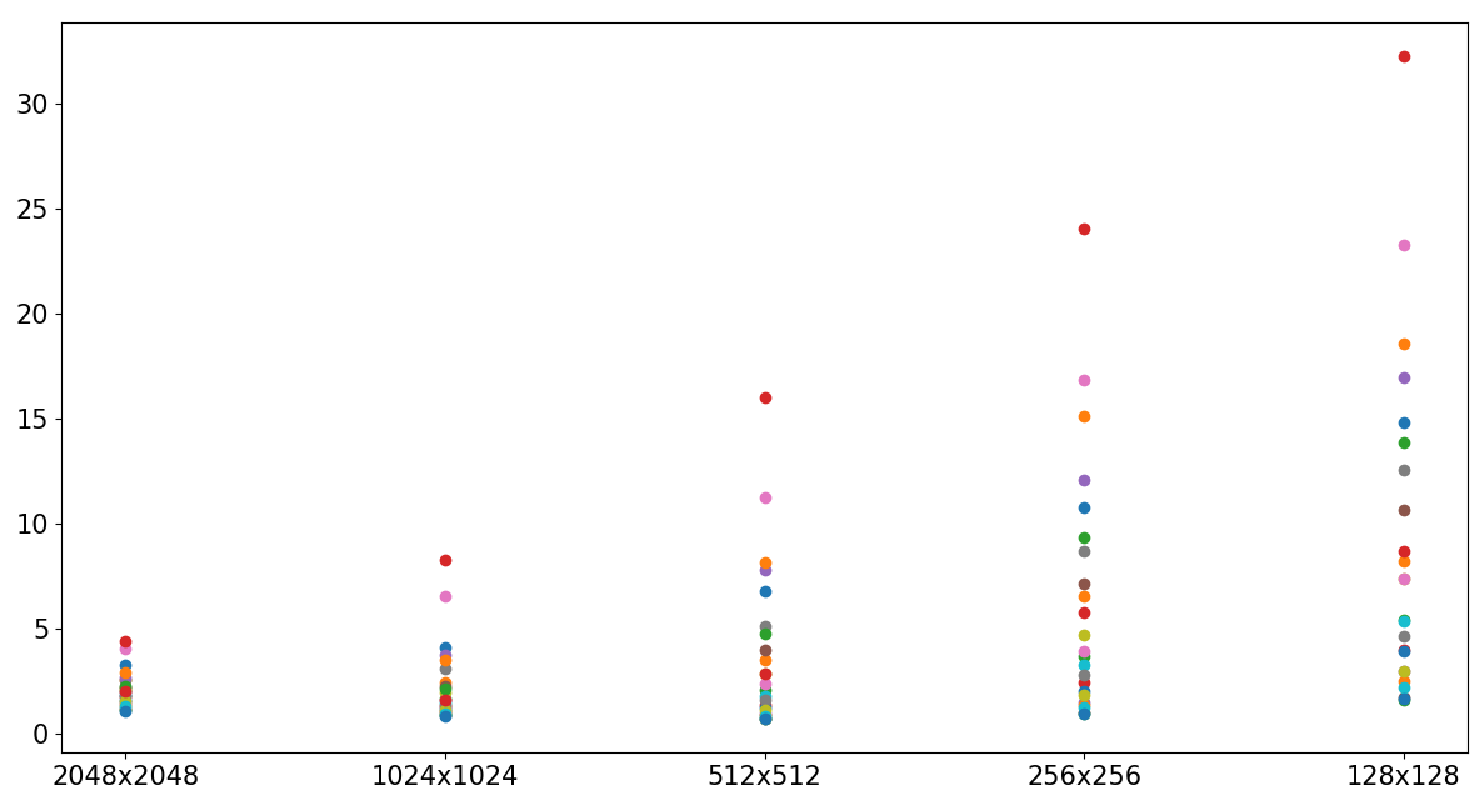, width = \linewidth}}
  \caption{Variation of $\beta$ coefficients values (axis Y) of a single image across multiple resolutions (axis X)}
  \label{fig:betas_plot}
 \end{figure}

The underlying hypothesis of our study is rooted in the observation that while the physical dimensions of an image can be altered through cropping, the intrinsic properties described by the $\beta$ coefficients of AC distributions remain consistent with the original resolution. These coefficients encapsulate critical information about the frequency content and texture details of the image, which are not significantly affected by not too big cropping operations. Therefore, by analyzing these $\beta$ coefficients, our model aims to identify the original resolution, independently of any cropping or resizing it may have undergone. The graph in Figure \ref{fig:betas_plot} depicts the trend of $\beta$ coefficient values across the same image at different resolutions, illustrating the distinct range of frequency components. By determining the likelihood that an image has been cropped without relying on metadata (which can be easily altered or removed), our model provides a computationally efficient and reliable method for detecting image manipulations. Such a tool would be invaluable in contexts where verifying the authenticity and integrity of digital images is crucial, in particular in digital forensics. Furthermore, our approach highlights the efficiency of using frequency domain features for image classification tasks. By relying on the $\beta$ values, the model leverages a compact yet informative representation of the image, facilitating rapid and resource-efficient processing. This aspect is particularly relevant in scenarios where computational resources are limited or when processing a large volume of images quickly is necessary.

The ML based classifiers play an important role in making sense of complex data, enabling computers to categorize or predict the group to which a new observation belongs based on a training dataset. A classifier algorithm sifts through data with known labels and learns from this data to predict the classification of unlabeled data. The performance and suitability of a classifier depend on the nature of the data and the specific task.

Among the most famous ML methods, many are used across a variety of applications: K-Nearest Neighbors (KNN), Decision Trees (in particular as Random Forests), Gradient Boosting Machines (GBM), and Support Vector Machines (SVM). All of those methods are able to give good results in general scenarios with some that work better in specific ones.  In order to understand what was the better classifier fitting our input data all $4$ the mentioned classifiers were tested with better results reached by the SVM. It was able to reach a general accuracy of $76\%$ (the specific discussion of the results is described in Section \ref{sec:results}), by achieving approximately $10\%$ more than Random forest and almost $20\%$ w.r.t. the others. SVM works by finding the hyperplane that best separates different classes in the feature space. The strength of SVMs lies in their use of kernels, which allow them to operate in a high-dimensional space, making them highly effective for non-linear data. The choice of the kernel function is critical, with the Radial Basis Function (RBF) kernel being particularly popular for its ability to handle non-linear relationships. The \textit{C} parameter trades off correct classification of training examples against maximization of the decision function's margin. The value chosen for our model is $100$: a high value of \textit{C} tells the model to give a higher priority to classifying all training examples correctly. Gamma hyper-parameter determines the influence of individual training examples. The low value used (0.1) suggests that each point has a moderate influence on the model's decision boundary. The choice of RBF kernel and the optimal values for \textit{C} and gamma were determined through Grid Search algorithm (using the python library scikit-learn), a method that performs exhaustive search over specified parameter values for an estimator. Grid Search evaluates and compares the performance of all possible combinations of parameter values, facilitating the selection of the best model. By leveraging the power of SVMs with carefully tuned hyperparameters, we aim to develop a highly accurate and reliable classifier capable of discerning the resolution of images, thereby contributing valuable insights to the field of digital image forensics.

\section{\uppercase{Results}}
\label{sec:results}

\begin{table*}[t]
	\centering
	\resizebox{0.7\linewidth}{!}{
	\renewcommand{\arraystretch}{2.9}
		\begin{tabular}{c||c||c||c||c||c}

			\hline \hline
		\diagbox[width=14em]{Real}{Predicted}&$2048\times 2048$ & $1024 \times 1024$ & $512 \times 512$ & $256 \times 256$ & $128 \times 128$  \\
			\hline \hline
                $2048 \times 2048$ & \cellcolor{blue!50}$98.43\%$  & $1.02\%$ & $0.14\%$ & $0.00\%$ & $0.41\%$  \\
			\hline
			$1024 \times 1024$ & $1.49\%$  & \cellcolor{blue!40}$87.79\%$ & $6.72\%$ & $1.76\%$ & $2.24\%$  \\
			\hline
			$512 \times 512$ & $1.02\%$  & $13.09\%$ & \cellcolor{blue!25}$64.76\%$ & $17.86\%$  & $3.27\%$   \\
			\hline
			$256 \times 256$& $0.43\%$  & $6.77\%$ & $27.93\%$ & \cellcolor{blue!20}$53.64\%$ & $11.23\%$ \\
                \hline
                $128 \times 128$ & $0.35\%$  & $4.01\%$ & $3.18\%$  & $15.57\%$ & \cellcolor{blue!35}$76.9\%$ \\
			\hline
		
		\end{tabular}
	}
	\caption{Confusion Matrix of the SVM Classifier}
	\label{tab:confusion_matrix}
\end{table*} 

The trained SVM model was tested with the $20\%$ of the dataset (about $7200$ non-cropped images) to determine its ability to correctly classify the input images. The model achieved an overall accuracy  of $76.55\%$, indicating an important level of precision for the task of resolution classification; in addition, the confusion matrix shown in Table \ref{tab:confusion_matrix}, describes the classification performance for each resolution. In future studies, we plan to investigate the anomaly observed in the accuracy growth for classifying $128\times128$ images, aiming to understand and optimize the underlying factors contributing to this trend.


\subsection{Cropping detection results}
\label{sec:crop_detection}
As explained in the previous sections our idea is to use the model (trained for resolution classification) to potentially detect cropping in an image. We start creating from RAISE $200$ images (not used during SVM training) of resolution $2048\times2048$ following the pipeline described in Section \ref{sec:dataset}, then, on each image $4$ different central cropping were carried out ($128\times128$, $256\times256$, $512\times512$ and $1024\times1024$), generating $800$ cropped images. To test the cropping detection of a dataset with starting resolution $2048\times2048$  the following strategy was employed: given a cropped image $C$, if the classifier predicts a resolution higher than actually size of $C$ a cropping was detected, otherwise no crop were detected. 
It is important to note that the test was performed in a so-called aligned scenario, i.e. when applying the crop, it was done respecting the $8\times8$ grid used to extract the DCT blocks; specifically, considering the image matrix during the crop, the number of rows deleted at the top and the number of columns deleted at the left is a multiple of $8$.
The accuracy of the model improved substantially with the increase in resolution of the cropped images. This improvement can be attributed to the retention of more image features that are representative of the original resolution, which enhances the classifier's ability to make correct predictions. 
The test carried out on the $800$ cropped images, in terms of resolution classification give us the following accuracies:
\begin{itemize}
  \item $128\times128$ cropped images: Accuracy = 76.0\%
  \item $256\times256$ cropped images: Accuracy = 82.5\%
  \item $512\times512$ cropped images: Accuracy = 89.5\%
  \item $1024\times1024$ cropped images: Accuracy = 99.0\%
\end{itemize}
where the $76\%$ shown for the images $128\times128$ means the percentage of time where the images were classified as $2048\times2048$. Employing the same results, to calculate the cropping detection using the strategy described above we obtained the results described in Table \ref{tab:cropping_results}; specifically, all the results are related to the resolution classification, only the last column is referred to the crop detection.

\begin{table*}[p]
	\centering
	\resizebox{0.8\linewidth}{!}{
	\renewcommand{\arraystretch}{3.5}
		\begin{tabular}{c||c||c||c||c||c||c}
			\hline \hline
		\diagbox[width=14em]{Cropping size}{Predicted  \\resolution}&$2048\times 2048$ & $1024 \times 1024$ & $512 \times 512$ & $256 \times 256$ & $128 \times 128$ & \% crop detection \\
			\hline \hline
			$1024 \times 1024$ & $99\%$  & $1\%$ & $0\%$ & $0\%$ & $0\%$ & \cellcolor{blue!45}$99\%$ \\
			\hline
			$512 \times 512$ & $89.5\%$  & $4\%$ & $1\%$ & $0.5\%$  & $5\%$ & \cellcolor{blue!35}$93.5\%$  \\
			\hline
			$256 \times 256$& $82.5\%$  & $4.5\%$ & $2\%$ & $0.5\%$ & $10.5\%$ &  \cellcolor{blue!25}$89\%$\\
                \hline
                $128 \times 128$ & $76\%$  & $3\%$ & $1.5\%$  & $3\%$ & $16.5\%$ & \cellcolor{blue!15}$83.5\%$ \\
			\hline
		
		\end{tabular}
	}
        \captionsetup{width=0.8\linewidth}
	\caption{Crop detection accuracy and classification results of images cropped from $2048\times2048$ resolution with cropping sizes equal to classification categories}
	\label{tab:cropping_results}
\end{table*}

\begin{table*}[p]
	\centering
	\resizebox{0.8\linewidth}{!}{
	\renewcommand{\arraystretch}{3.5}
		\begin{tabular}{c||c||c||c||c||c||c}

			\hline \hline
		\diagbox[width=14em]{Cropping size}{Predicted  \\resolution}&$2048\times 2048$ & $1024 \times 1024$ & $512 \times 512$ & $256 \times 256$ & $128 \times 128$ & \% crop detection \\
			\hline \hline
                $1536 \times 1536$& $100\%$  & $0\%$ & $0\%$ & $0\%$ & $0\%$ &  \cellcolor{blue!50}$100\%$\\
                \hline
                $750 \times 750$ & $96\%$  & $2\%$ & $0\%$ & $0\%$  & $2\%$ & \cellcolor{blue!48}$98\%$  \\
			\hline
			$350 \times 350$ & $85.5\%$  & $4.5\%$ & $1.5\%$ & $1\%$ & $7.5\%$ & \cellcolor{blue!40}$91.5\%$ \\
			\hline	
		\end{tabular}
	}
        \captionsetup{width=0.8\linewidth}
	\caption{Crop detection accuracy and classification results of images cropped from $2048\times2048$ resolution with cropping sizes different w.r.t classification categories}
	\label{tab:cropping_results_2}
\end{table*} 

\begin{table*}[p]
	\centering
	\resizebox{0.8\linewidth}{!}{
	\renewcommand{\arraystretch}{3.5}
		\begin{tabular}{c||c||c||c||c||c||c}
			\hline \hline
		\diagbox[width=14em]{Cropping size}{Predicted  \\resolution}&$2048\times 2048$ & $1024 \times 1024$ & $512 \times 512$ & $256 \times 256$ & $128 \times 128$ & \% crop detection \\
			\hline \hline
                $512 \times 512$& $3.5\%$  & $76.5\%$ & $13\%$ & $1.5\%$ & $5.5\%$ &  \cellcolor{blue!15}$80\%$\\
                \hline
                $256 \times 256$ & $3.5\%$  & $63.5\%$ & $13.5\%$ & $8\%$  & $11.5\%$ & \cellcolor{blue!15}$80.5\%$  \\
			\hline
			$128 \times 128$ & $4\%$  & $47\%$ & $6\%$ & $11\%$ & $32\%$ & \cellcolor{blue!10}$68\%$ \\
			\hline	
		\end{tabular}
	}
        \captionsetup{width=0.8\linewidth}
	\caption{Crop detection accuracy and classification results of images cropped from $1024\times1024$ resolution with cropping sizes different w.r.t classification categories}
	\label{tab:cropping_results_1024_1024}
\end{table*} 

\subsection{Generalization}

In addition to the primary experiments, additional tests were performed to evaluate the model with images cropped at different sizes w.r.t. the classes used to train the classifier. Then, using the same cropping pipeline described in Section \ref{sec:crop_detection}, another $600$ cropped images were generated. Specifically $3$ cropping sizes ($1536\times1536$, $750\times750$, $350\times350$) were used, representing an intermediate value between the resolutions employed in the training phase. The results in Table \ref{tab:cropping_results_2} demonstrate how our pipeline to detect the crop, is not dependent w.r.t. the cropping size; moreover, the detection accuracy in Table \ref{tab:cropping_results_2} fits perfectly the trend of the previous one (Table \ref{tab:cropping_results}). The trend of all discussed results shows how they are strictly dependent on the resolutions of the images; this turns out to be easy to understand because the more information and its variability, the easier it is to perform detection and recognition tasks. To complete our study, the same test performed in Section \ref{sec:crop_detection} was carried out starting from images with $1024\times1024$ resolution. Table \ref{tab:cropping_results_1024_1024} shows the classifier results, also in terms of cropping detection, confirming what was discussed before; nevertheless, the accuracies related to the cropping detection demonstrate the usefulness of the features on this task and that future research in this direction will definitely lead to important results.



\section{\uppercase{Conclusions}}
\label{sec:conclusion}

This paper presents a preliminary study on $\beta$ values of AC distributions for cropping detection. A classifier was developed to detect the resolution of images between some classes. After the application of a central cropping, we tested how classifier can accurately detect the image's native resolution. Finally, through a proper strategy for crop detection, we demonstrated how the classifier could be employed for cropping detection, confirming the information contained in $\beta$ values of AC distributions. The proposed method is limited by its categorization into only five resolution classes within the SVM framework. Future work could involve refining the SVM by searching for more optimal hyperparameters. Continual tuning of these parameters could yield a model that performs with even greater precision. To improve the robustness and versatility of the classifier, we plan to train it on a more comprehensive dataset that encompasses a wider range of image resolutions, adding more resolution classes to the model. Deep learning approaches were not incorporated at this stage due to the requirement for a more extensive and heterogeneous dataset beyond what is available in RAISE. The use Convolutional Neural Network (CNN) may provide better performances in this tasks due to their hierarchical feature extraction capabilities, permitting us to investigate challenging scenarios such as lower resolutions, non-aligned crops or compressed images.

\section*{\uppercase{Acknowledgements}}

The work of Claudio Vittorio Ragaglia has been supported by the Spoke 1 "Future HPC \& BigData”  of the Italian Research Center on High-Performance Computing, Big Data and Quantum Computing (ICSC)  funded by MUR Missione 4 Componente 2 Investimento 1.4: Potenziamento strutture di ricerca e creazione di "campioni nazionali di R\&S (M4C2-19)" - Next Generation EU (NGEU). The work of Francesco Guarnera has been supported by MUR in the framework of PNRR PE0000013, under project “Future Artificial Intelligence Research – FAIR".

\bibliographystyle{apalike}
{\small
\bibliography{example}}

\begin{thebibliography}{}

\bibitem[Barni et~al., 2017]{barni2017aligned}
Barni, M., Bondi, L., Bonettini, N., Bestagini, P., Costanzo, A., Maggini, M., Tondi, B., and Tubaro, S. (2017).
\newblock Aligned and non-aligned double jpeg detection using convolutional neural networks.
\newblock {\em Journal of Visual Communication and Image Representation}, 49:153--163.

\bibitem[Battiato et~al., 2022]{battiato2022cnn}
Battiato, S., Giudice, O., Guarnera, F., and Puglisi, G. (2022).
\newblock Cnn-based first quantization estimation of double compressed jpeg images.
\newblock {\em Journal of Visual Communication and Image Representation}, 89:103635.

\bibitem[Battiato et~al., 2001]{battiato2001psychovisual}
Battiato, S., Mancuso, M., Bosco, A., and Guarnera, M. (2001).
\newblock Psychovisual and statistical optimization of quantization tables for dct compression engines.
\newblock In {\em Proceedings 11th International Conference on Image Analysis and Processing}, pages 602--606. IEEE.

\bibitem[Battiato and Messina, 2009]{battiato2009digital}
Battiato, S. and Messina, G. (2009).
\newblock Digital forgery estimation into dct domain: a critical analysis.
\newblock In {\em Proceedings of the First ACM workshop on Multimedia in forensics}, pages 37--42.

\bibitem[Dang-Nguyen et~al., 2015]{dang2015raise}
Dang-Nguyen, D.-T., Pasquini, C., Conotter, V., and Boato, G. (2015).
\newblock Raise: A raw images dataset for digital image forensics.
\newblock In {\em Proceedings of the 6th ACM multimedia systems conference}, pages 219--224.

\bibitem[Farid, 2009]{4806202}
Farid, H. (2009).
\newblock Image forgery detection.
\newblock {\em IEEE Signal Processing Magazine}, 26(2):16--25.

\bibitem[Galvan et~al., 2014]{galvan2014first}
Galvan, F., Puglisi, G., Bruna, A.~R., and Battiato, S. (2014).
\newblock First quantization matrix estimation from double compressed jpeg images.
\newblock {\em IEEE Transactions on Information Forensics and Security}, 9(8):1299--1310.

\bibitem[Giudice et~al., 2019]{10.1007/978-3-030-30645-8_65}
Giudice, O., Guarnera, F., Paratore, A., and Battiato, S. (2019).
\newblock 1-d dct domain analysis for jpeg double compression detection.
\newblock In Ricci, E., Rota~Bul{\`o}, S., Snoek, C., Lanz, O., Messelodi, S., and Sebe, N., editors, {\em Image Analysis and Processing -- ICIAP 2019}, pages 716--726, Cham. Springer International Publishing.

\bibitem[Giudice et~al., 2021]{giudice2021fighting}
Giudice, O., Guarnera, L., and Battiato, S. (2021).
\newblock Fighting deepfakes by detecting gan dct anomalies.
\newblock {\em Journal of Imaging}, 7(8):128.

\bibitem[Hou et~al., 2013]{hou2013double}
Hou, W., Ji, Z., Jin, X., and Li, X. (2013).
\newblock Double jpeg compression detection based on extended first digit features of dct coefficients.
\newblock {\em International Journal of Information and Education Technology}, 3(5):512.

\bibitem[Lam and Goodman, 2000]{LamGoodman2000}
Lam, E. and Goodman, J. (2000).
\newblock A mathematical analysis of the dct coefficient distributions for images.
\newblock {\em Journal of Image Processing}, 9(10):1661--1666.

\bibitem[Piva, 2013]{piva2013overview}
Piva, A. (2013).
\newblock An overview on image forensics.
\newblock {\em International Scholarly Research Notices}, 2013.

\bibitem[Rav{\`\i} et~al., 2016]{ravi2016semantic}
Rav{\`\i}, D., Bober, M., Farinella, G.~M., Guarnera, M., and Battiato, S. (2016).
\newblock Semantic segmentation of images exploiting dct based features and random forest.
\newblock {\em Pattern Recognition}, 52:260--273.

\bibitem[Tondi et~al., 2021]{tondi2021boosting}
Tondi, B., Costanzo, A., Huang, D., and Li, B. (2021).
\newblock Boosting cnn-based primary quantization matrix estimation of double jpeg images via a classification-like architecture.
\newblock {\em EURASIP Journal on Information Security}, 2021(1):5.

\bibitem[Uricchio et~al., 2017]{uricchio2017localization}
Uricchio, T., Ballan, L., Roberto~Caldelli, I., et~al. (2017).
\newblock Localization of jpeg double compression through multi-domain convolutional neural networks.
\newblock In {\em Proceedings of the IEEE Conference on Computer Vision and Pattern Recognition Workshops}, pages 53--59.

\end{thebibliography}

\end{document}